\pdfoutput=1

\documentclass[11pt]{article}
\usepackage{amsmath}
\usepackage{amsfonts}
\usepackage{microtype}
\usepackage{multirow}
\usepackage{multicol}
\usepackage{booktabs}
\usepackage{threeparttable}
\usepackage{graphicx}
\usepackage{epstopdf}
\usepackage{color}
\usepackage{enumitem}
\usepackage{bm}
\usepackage{amssymb}
\usepackage{ragged2e}
\usepackage{emnlp2021}
\usepackage{makecell}
\usepackage{cleveref}
\crefname{section}{§}{§§}

\usepackage{times}
\usepackage{latexsym}

\usepackage[T1]{fontenc}

\usepackage[utf8]{inputenc}

\usepackage{microtype}

\title{Topic-Guided Abstractive Multi-Document Summarization}

\author{Peng Cui \and Le Hu\\
School of Computer Science and Technology\\
Harbin Institute of Technology, Harbin, China\\
\texttt{\{pcui, lhu\}@insun.hit.edu.cn}}

\begin{document}
\maketitle
\begin{abstract}
  A critical point of multi-document summarization (MDS) is to learn the relations among various documents. In this paper, we propose a novel abstractive MDS model, in which we represent multiple documents as a heterogeneous graph, taking semantic nodes of different granularities into account, and then apply a graph-to-sequence framework to generate summaries. Moreover, we employ a neural topic model to jointly discover latent topics that can act as cross-document semantic units to bridge different documents and provide global information to guide the summary generation. Since topic extraction can be viewed as a special type of summarization that “summarizes” texts into a more abstract format, i.e., a topic distribution, we adopt a multi-task learning strategy to jointly train the topic and summarization module, allowing the promotion of each other. Experimental results on the Multi-News dataset demonstrate that our model outperforms previous state-of-the-art MDS models on both Rouge metrics and human evaluation, meanwhile learns high-quality topics. 
\end{abstract}

\section{Introduction}

Multi-document summarization (MDS) is the task to create a fluent and concise summary for a collection of thematically related documents. Compared to single document summarization, it requires the ability to incorporate the perspective from multiple sources and therefore is arguably more challenging \citep{lin2019abstractive}. Broadly, existing studies can be classified into two categories: extractive and abstractive. Extractive approaches directly select important sentences from the input documents, which is usually regarded as a sentence labeling \cite{nallapati2016summarunner,zhang2018neural,dong2018banditsum} or sentence ranking task \citep{narayan2018don}. By contrast, abstractive models typically use the natural language generation technology to produce a word-by-word summary. In general, extractive methods are more efficient and can avoid grammatical errors \citep{cui2020enhancing}, while abstractive methods are more flexible and human-like because they can generate absent words\citep{lin2019abstractive}.

Recently, with the development of representation learning for NLP \citep{vaswani2017attention,devlin2018bert} and large-scale datasets \citep{fabbri2019multi}, some studies have achieved promising results on abstractive MDS \citep{liu2019hierarchical,jin2020multi}. Nevertheless, we found there are two limitations that have not been addressed by previous studies. First, some works simply concatenate multiple documents into a flat sequence and then apply single-document summarization approaches \citep{liu2018generating,fabbri2019multi}. However, this paradigm fails to consider the hierarchical document structures, which plays a key role in MDS task \citep{jin2020multi}. Also, the concatenation operation inevitably produces a lengthy sequence, and encoding long texts for summarization is a challenge \citep{cohan2018a}.

Second, when dealing with multiple documents, a critical point is to learn the cross-document relations. Some studies address this problem by mining the co-occurrence words or entities \citep{wang2020heterogeneous}, which can hardly capture implicit associations due to the diverse language expressions. Some other studies \citep{jin2020multi,liu2019hierarchical} first generate low-dimensional vectors in sentence- or paragraph-level and then build interaction based on these highly compressed representations. These methods inevitably result in the loss of large amounts of fine-grained interaction features and would damage the interpretability of models. Therefore, how to learn the relation across documents effectively remains an open question.

To shed lights on these missing points, this paper proposes a novel abstractive MDS model that marries topic modeling into abstractive summary generation. The motivation is that both tasks aim to distil salient information from massive text and therefore could provide complementary features for each other.
Concretely, we jointly optimize a \emph{neural topic model} (NTM) \citep{miao2017discovering,srivastava2017autoencoding} that learns topic distribution of source documents and corpus-level topic representations, and an \emph{abstractive summarizer} that incorporates latent topics to summary generation process. 
In the encoding process, we represent multiple documents as a heterogeneous graph consisting of word, topic, and document nodes and encode it with a graph neural network to capture the interactions among different semantic units. In the decoding process, we devise a topic-aware decoder that leverages learned topics to guide the summary generation. 
We train the two modules with a multi-task learning framework, where an \emph{inconsistency loss} is applied to penalize the difference between the topic distribution of source documents and that of generated summaries. It encourages the summarizer to generate a summary that is thematically consistent with its source documents and also helps the two modules learn from each other. In this manner, our model is learned such that better topics can yield better summaries and vice versa.

We conduct throughout experiments on the recently released Multi-News dataset \citep{fabbri2019multi}. The results demonstrate the effectiveness and superiority of our model.
To sum up, the contributions of this paper are threefolds:

1) To the best of our knowledge, we carry out the first systematic study on jointly modeling topic inference and abstractive MDS and demonstrate the positive mutual effect between the two tasks.

2) We propose a novel MDS model that joint optimizes a neural topic model and an abstractive summarizer. We propose an inconsistency loss to penalize the disagreement between the two modules and help them learn from each other.

3) Experimental results on the Multi-News dataset demonstrate that our model achieves the state-of-the-art performance on both Rouge scores and human evaluation, meanwhile learns high-quality topics.
\section{Related Work}
\noindent \textbf{Multi-document summarization} is a challenging subtask of text summarization with a long history. Many previous methods are extractive partly due to the lack of sufficient training data. These methods usually compute sentence salience over graph structures \citep{mihalcea2004textrank,wang2020heterogeneous}.
Abstractive MDS studies have been fueled by the recent development of large-scale datasets \citep{fabbri2019multi} and representation learning of NLP \citep{vaswani2017attention}. Among them, hierarchical networks \citep{liu2019hierarchical} and graph neural networks \citep{jin2020multi} are widely used to capture the cross-document relations. However, most of them build interaction based on word- or paragraph-level representations, which are not flexible or straightforward. In comparison, we propose to model multiple documents more effectively by mining their subtopics.

\noindent \textbf{Datasets for multi-document summarization} Recently, \citet{fabbri2019multi} released the first large-scale news dataset for MDS. Each article is collected from real-life scenarios and the golden summaries are written by human, which ensures the data quality. Prior to them, some studies tried to construct dataset in automatic manners. For example, \citet{liu2018generating} and \citet{liu2019hierarchical} built datasets based on Wikipedia pages, regarding the first section as the summary and others as different documents. However, modeling the relations among different documents is a different task from modeling that of paragraphs from a same document. 
Therefore, the generalization ability of models built on such data could be questionable. 
For this reason, we do not consider such auto-constructed datasets but focus on the Multi-News dataset curated by human.

\noindent \textbf{Topic modeling for text summarization} Topic model is widely used for document modeling. Nevertheless, few studies have applied it in summarization task. Previous studies regarded topical distributions as additional features to enrich word or sentence representations \cite{wei2012document,narayan2018don,wang2020friendly}.  
However, these methods use a pipeline process where topic extraction and summary generation are separately performed. In comparison, we adopt a multi-task learning strategy so that the two tasks can learn complementary features from each other. Recently, \citet{cui2020enhancing} has applied NTM to extractive summarization. Though inspired by it, the motivation and proposed method of this study differ from it by a large margin. \citet{cui2020enhancing} use latent topics to preselect salient sentences, while we use them to capture cross-document relations for abstractive MDS. Besides, \citet{cui2020enhancing} solely explores the effect of topic modeling on summarization, while we systematically explore their interplay.

\begin{figure*}[tp]
  \centering
  \includegraphics[scale=0.55]{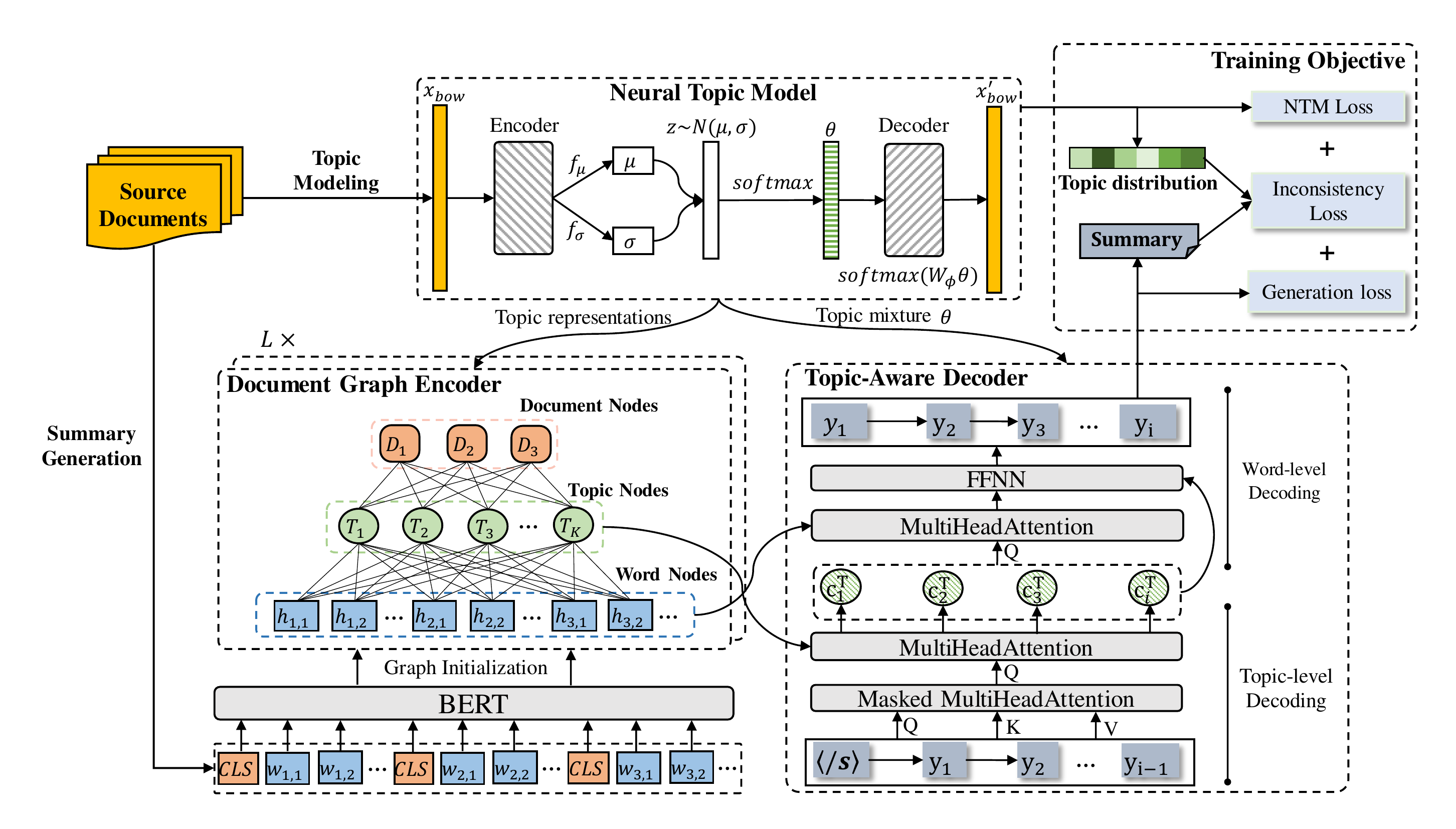}
  \caption{An illustration of the proposed model (\emph{TG-MultiSum}). The summarizer consists of a \textbf{document graph encoder} (left bottom) to encode the source documents and a \textbf{topic-aware decoder} (bottom right) to generate summary words. A \textbf{neural topic model} (top) is applied to provide topical information as guidance signals. The two parts are jointly trained with an \emph{inconsistency loss} to penalize their disagreement.}
  \label{fig:model}
\end{figure*}

\section{Model}\label{sec:3}
This section describes our model, named as topic-guided multi-document summarization (\emph{TG-MultiSum}). The overall architecture is presented in Figure \ref{fig:model}. Given a set of documents $\{D_1, D_2, ..., D_N\}$, the goal of our model is to generate a word sequence $S=\{y_1, y_2, ..., y_s\}$ as the summary. Our model consists of three major components: 1) the \textbf{neural topic model} aims to learn the topical information of source documents; 2) the \textbf{document graph encoder} builds the interaction among different documents and various semantic units. 3) the \textbf{topic-aware decoder} generates summary words based on the learned node representations. The entire model is trained in an end-to-end manner. We explain each part below.
\subsection{Neural Topic Model}\label{sec:3.1}
One innovation of this study is that it incorporates topical information into summarization explicitly. Based on the current development of topic modeling, we employ a VAE-based neural topic model proposed in \citet{miao2017discovering} to discover latent topics. Compared with conventional LDA-style topic models, it can be trained together with neural networks and therefore has better adaptability \citep{zeng2018topic,cui2020enhancing}.

Similar to LDA, NTM assumes the existence of $K$ underlying topics throughout the corpus. Each document can be represented as a $K$-dim topic mixture, and each topic can be represented as a distribution over the vocabulary. NTM learns the topics through an encoding–decoding process. 
Let $x_{bow}\in \mathbb{R}^{|V|} $ denote the bag-of-word term vector of input documents, where $|V|$ is the vocabulary size.
We first use an MLP encoder to estimate its exclusive priors $\sigma$ and $\mu $, which are used to generate the topic distribution through a Gaussian softmax, as shown in the following:

\begin{align}
    \sigma = f_{\sigma}(x_{bow}), \mu &= f_{\mu}(x_{bow}), \label{eq:1} \\ 
    z \sim N(\sigma, \mu^{2}), \theta_{x}&=softmax(z), \label{eq:2}
\end{align}
where $f_{*}(\cdot)$ is a neural perceptron with ReLU activation. $z,\theta_{x} \in \mathbb{R}^K$ are the latent variable and topic distribution of input documents, respectively.

Then, we use a $softmax$ layer to reconstruct the input text, i.e., $x_{bow}^{'}=softmax({\rm W_{\phi}}\theta_{x})$. In particular, the weight matrix ${\rm W_{\phi}} \in \mathbb{R}^{|V|\times K}$ can be regarded as the unnormalized topic–word distributions, where ${\rm W_{\phi}}^{i,j}$ indicates the relevance between the $i$-th word and $j$-th topic.

\subsection{Document Graph Encoder}\label{sec:3.2}
\textbf{Graph Construction} \quad Inspired by the assumption of LDA, we view the input documents as a three-layer graph consisting of document, topic, and word units. Formally, let $\mathcal{G} =(V,E)$ denote the constructed graph. The node set $V$ consists of $N$ document nodes $\{v_{1}^{d},v_{2}^{d},...,v_{N}^{d}\}$, $K$ topic nodes $\{v_{1}^{t},...,v_{K}^{t}\}$, and $M$ word nodes $\{v_1^w,...,v_M^w\}$. The edge set is defined as $E=E_{D,T}\cup E_{T,W}$, where $E_{D,T}=\{e_{1,1},...,e_{N,K}\}$ represents the document–topic edges, and $E_{D,T}=\{e_{1,1},...,e_{K,M}\}$ represents the topic–word edges.

\noindent \textbf{Graph Initialization} \quad
To capture the contextual information, we use a shared BERT encoder to encode each document independently, which has been proved effective in summarization task.
The output states of each word $\bm{{\rm H^W}}=\{ h_{1}^{1},...,h_{|D_1|}^{1},...,h_{1}^{N},...,h_{|D_N|}^{N}\}$ are used as the initial word node representations, and those of ${\rm [CLS]}$ tokens $\bm{{\rm H^D}}=\{h_{CLS}^{1},...,h_{CLS}^{N} \}$ are used as the initial document node representations.   

As for the topic nodes, we use the weight matrix ${\rm W_{\phi}}$ learned from NTM as raw features and transform it to low-dim topic representations via $\bm{{\rm H^T}} = f_{\phi}(W_{\phi})$,
where $f_{\phi}(\cdot)$ is a Tanh-activated neural perceptron. Each row of $\bm{{\rm H^T}} \in \mathbb{R}^{K\times d}$ is a topic vector with predefined dimension $d$.

\noindent \textbf{Graph Propagation} \quad Given the constructed graph and its initial node representations, we then use a graph neural network to capture the relations among different semantic units. Here we present a single \emph{Document Graph Encoder} (DGE) layer. Multiple DGE layers are stacked in our experiments.

Let $u_{i}^{l}$ be the $i$-th node representation in the $l$-th layer. The updating process of $u_{i}^{(i)}$ is denoted as follows:
\begin{align}
  \tilde{u}_{i}^{l} =& {\rm W_{1}^{l}} {\rm Relu}({\rm W_{2}^{l}} u_{i}^{l} + {\rm b_{1}^{l}}) + {\rm b_{2}^{l}}, \label{eq:3} \\
  z_{i,j}^{l} =& {\rm LeaklyReLU} ({\rm W_3^{l}} [f_{t}(\tilde{u}_{i}^{l});f_{t}(\tilde{u}_{j}^{l})]), \label{eq:4} \\
  \alpha_{i,j}^{l} =& \frac{exp(z_{i,j}^{l})}{\sum_{k\in\mathcal{N}_{i}}exp(z_{i,k}^{l})}, \label{eq:5} \\
  \vec{u}_i^{l} =& ||_{m=1}^{M} \sum_{j\in N_i}tanh(\alpha_{i,j}^{l,m} {\rm W_4^m} \vec{u}_j^{l}), \label{eq:6}
\end{align}
where ${\rm W_*}$ and ${\rm b_*}$ denote trainable parameters. Eqs.\ref{eq:4}–\ref{eq:6} are the graph attention network (GAT)\citep{velickovic2018graph} which updates each node by aggregating its neighbor nodes $\mathcal{N}_{*}$. 

Note that the vanilla graph attention network is designed for homogeneous graphs. However, in our task, word, document, and topic nodes should be considered different semantic units. Therefore, we make a modification in Eq.\ref{eq:4} by adding a node-type function $f_{t}(\cdot)$. It uses exclusive parameters for different node types to project them into a common vector space where the attention score is calculated.

To construct deep networks, we further add a residual connection and layer normalization \citet{ba2016layer} to connect adjacent DGE layers.
\begin{gather}
  u_{i}^{l+1} = {\rm LayerNorm}(u_{i}^{l} + {\rm Dropout}(\vec{u}_{i}^{l})).
\end{gather}

\subsection{Topic-Aware Decoder}\label{sec:3.3}
To better utilize the guidance effect of latent topics, our decoder, being topic-aware, adopts a two-step decoding process. In each step, it first decodes the current topic, and then generates summary words correspondingly.

\noindent \textbf{Topic-level decoding} \quad The topic context $c_i^T$ in $i$-th step is conditioned on the previous decoded words $\mathcal{Y}_{<i}=\{y_1, y_2, ..., y_{i-1}\}$ and the topic representations $\bm{{\rm H^T}}$ output from the graph encoder, as shown as follows:
\begin{align}
  u_{i-1} =& {\rm MHAttn}(e_{i-1},\bm{{\rm E_{<i-1}}},\bm{{\rm E_{<i-1}}}), \label{eq:8} \\
  c_{i}^{T} =& {\rm MHAttn}(u_{i},\bm{{\rm H_{T}}},\bm{{\rm H_{T}}}), \label{eq:9}
\end{align}
where ${\rm MHAttn}({\rm Q},{\rm K},{\rm V})$ denotes the multi-head attention introduced in \citet{vaswani2017attention}. The first attention layer is used to capture contextual feature of decoded sequence, while the second is to incorporate topical information.

In effect, $c_{i}^{T}$ can be viewed as a \emph{topic pointer} that indicates which topics should be discussed in the current step.

\noindent \textbf{Word-level decoding} \quad We then use the generated $c_{i}^{T}$ to guide the word prediction. Another ${\rm MHAttn}$ layer is first applied to select relevant parts of source word sequence $\bm{{\rm H^W}}$ with $c_{i}^{T}$ as the query, followed by a neural perceptron to inject the current topic focus.
\begin{gather}
  v_{i} = {\rm MHAttn}(c_{i}^{T},\bm{{\rm H^{W}}}, \bm{{\rm H^{W}}}),\\
  o_{i} = {\rm tanh}({\rm W_o}[v_{i};c_{i}^{T}]+{\rm b_o}),
\end{gather}
where $o_{i}$ is the final output representation in $i$-th step of the decoder.  

The predicted word distribution over the vocabulary is computed through a softmax layer, i.e., $p_{i}^{g}=softmax({\rm W_g} o_{i-1}+{\rm b_g})$. To alleviate the out-of-vocabulary (OOV) problem, we employ the copy mechanism \citep{see2017get} to allow the generator to copy words from source documents. The copy distribution $p_{i}^{c}$ is computed as follows.
\begin{align}
  \varepsilon_t =& softmax(\bm{{\rm H^{W}}} o_{t}),\\
  p_{t}^{c} =& \sum_{i\leq N} \sum_{j \leq |D_{i}|} \varepsilon_t z_{i,j},  
\end{align}
where $\varepsilon_t$ is the attention weight of source words, and $z_{i, j}$ is the one-hot indicator vector for word $w_{i,j}$. The final generation distribution is the linear combination of $p_{i}^{g}$ and $p_{i}^{c}$, as shown as follows:
\begin{gather}
  p_i=\eta_{i}\ast p_{i}^{c}+ (1-\eta_{i}) \ast  p_{i}^{g}, \\
  \eta_{i}=\sigma({\rm W_{\eta}} o_{i} +{\rm b_{\eta}}),
\end{gather}
where $\sigma$ indicates sigmoid function and $\eta_{i}$ is the copy weight.
\subsection{Joint Learning with Inconsistency Loss}\label{subsection:3.4}
Since text summarization and topic modeling both aim to distill salient information from input documents, we jointly train the two modules to help them learn complementary information from each other. 
The loss function of our model consists of three parts. 
The summary generation loss $\mathcal{L}_{gen}$ is defined as the negative log-likelihood of ground-truth words, i.e.,
\begin{gather}
  \mathcal{L}_{gen}=\sum_{c\in\mathcal{C}}\sum_{w\in y^c}log p(w).
\end{gather}

The NTM loss $\mathcal{L}_{NTM}$ is based on the evidence lower bound, i.e., 
\begin{gather}
  \mathcal{L}_{NTM}={\rm KL}(p(z)\parallel q(z|x)) - \mathbb{E}_{q(z)}[p(x|z)],
\end{gather}
where the first term is the ${\rm KL}$ divergence, and the second term indicates the construction loss. $p(\cdot)$ and $q(\cdot)$ are the encoder and decoder networks described in \cref{sec:3.1}, respectively. 

We also devise an inconsistency loss $\mathcal{L}_{inc}$ to penalize the disagreement between the topic distribution of generated summary and that of source documents, as shown as follows:
\begin{gather}
  \mathcal{L}_{inc} = {\rm KL}(\theta_{x}|| \sum_{1 \leq  i \leq L} \theta_{dec}^{i}), \label{eq:18}
\end{gather}
where $\theta_{x}$ is the document topic mixture learned from NTM(Eq.\ref{eq:2}), and $\theta_{dec}^{i}$ is the topic distribution of $i$-th decoding step computed in Eq.\ref{eq:9}.

The final loss is the linear combination of the three parts, i.e., $L=\mathcal{L}_{gen} + \gamma*\mathcal{L}_{NTM}+ \tau*\mathcal{L}_{inc}$, where $\gamma$ and $\tau$ are hyperparameters.

\section{Experimental Setup}
\subsection{Dataset}\label{sec:4.1}
We conduct experiments on the recently constructed dataset Multi-News \citep{fabbri2019multi}. The standard split contains 44972/5622/5622 instances for training, validation, and test. Each instance consists of a set of news articles paired with a human-written summary. 
The average summary length and article cluster length are 264 and 2103, respectively.
In Table \ref{tab:distribution-of-number}, we present the distribution of the number of source articles per summary. As shown, nearly half of the summaries are paired with at least three source articles, which highly demands the ability to process multi-source information. 
The average input length (2103) also brings difficulty for the encoder network.
These characteristics make the dataset a good challenge for the MDS task.
\begin{table}[t]
  \begin{center}
  \scalebox{0.8}{
    \begin{tabular}{cll|cll}
    \toprule 
    \bf \# of source  & \bf Freq & \bf Prop  & \bf \# of source  & \bf Freq & \bf Prop \\ 
    \hline
      2 & 23,894 & 53.1\% & 7  & 382 & 0.8\% \\
      3 & 12,707 & 28.3\% & 8  & 209 & 0.5\% \\
      4 & 5,022  & 11.2\% & 9  &  89 & 0.2\% \\
      5 & 1,873  &  4.2\% & 10 &  33 & 0.1\% \\
      6 & 763    &  1.7\% &    &     &       \\
    \bottomrule
    \end{tabular}}
  \end{center}
  \caption{\label{tab:distribution-of-number} The distribution of the number of source documents in the Multi-News dataset.}
\end{table}
\subsection{State-of-the-art Baselines}\label{sec:4.2}
We compare our model with the state-of-the-art extractive and abstractive models. The abstractive baselines are as follows.

\noindent \underline{\textbf{PGN}} \cite{see2017get}, pointer-generator network extends the standard seq2seq framework with copy and coverage mechanism.

\noindent \underline{\textbf{Hi-MAP}} \citep{fabbri2019multi} extends PGN into a hierarchical structure and integrates a MMR module to minimize redundancy.

\noindent \underline{\textbf{CopyTransformer}} \citep{gehrmann2018bottom} randomly chooses one of the attention heads of Transformer as the copy distribution.

\noindent \underline{\textbf{MGSum-abs}} \citep{jin2020multi} is a state-of-the-art abstractive MDS model. It designs an interaction network to integrate information from different granularities.

We also compare with the following extractive baselines:

\noindent \underline{\textbf{HiBERT}} \citep{zhang2019hibert} modifies the standard BERT to a hierarchical structure. We migrate it to MDS by concatenating the input documents.

\noindent \underline{\textbf{Hi-Transformer}}\citep{liu2019hierarchical} adds additional attention heads to the Transformer to share the information across documents.

\noindent \underline{\textbf{HeterGrapSum}} \citep{wang2020heterogeneous} uses a heterogeneous graph neural network to encode word, sentence, and document nodes.

\noindent \underline{\textbf{MatchSum}} \citep{2020Extractive} regards content selection as a text matching problem. It has reported the state-of-the-art results on Multi-News dataset.

\noindent \underline{\textbf{MGSum-ext}} \citep{jin2020multi} is the extractive version of MGSum-abs.

\begin{table}[tp]
  \small
  \centering
  \begin{tabular}{c|ccc}
  \toprule[1pt]
  \textbf{model}   & \textbf{R-1}   & \textbf{R-2}   & \textbf{R-SU}  \\ \hline
  \multicolumn{4}{c}{\textit{Non-Neural methods}}                     \\ \hline
  Lead-3$^{\dagger}$           & 39.41          & 11.77          & 14.51          \\ 
  LexRank$^{\dagger}$          & 38.27          & 12.70           & 13.20           \\ 
  TextRank$^{\dagger}$         & 38.44          & 13.10           & 13.50           \\ 
  MMR$^{\dagger}$              & 38.77          & 11.98          & 12.91          \\ \hline
  \multicolumn{4}{c}{\textit{Neural-based   Extractive Models}}       \\ \hline
  Hi-BERT          & 43.86          & 14.62          & 18.34          \\ 
  MGSum-ext        & 44.75          & 15.75          & 19.30           \\ 
  HeterGraphSum    & 46.05               & 16.35               & 17.81               \\ 
  MatchSum         & 46.20           & 16.51          & 20.05          \\ \hline
  \multicolumn{4}{c}{\textit{Abstractive Models}}                     \\ \hline
  PGN              & 41.85          & 12.91          & 16.46          \\ 
  Hi-Map           & 43.47          & 14.89          & 17.41          \\ 
  Copy Transformer & 43.57          & 14.03          & 17.37          \\ 
  Hi-Transformer   & 43.85          & 15.60           & 18.80           \\ 
  MGSum-abs        & 46.00             & 16.81          & 20.09          \\ \hline
  \multicolumn{4}{c}{\textit{Ours}}                     \\ \hline
  TG-MultiSum${\rm _{pip}}$   & 46.04          & 16.43          & 19.82          \\ 
  TG-MultiSum    & \textbf{47.10} & \textbf{17.55} & \textbf{20.73} \\
  \bottomrule[1pt]  
  \end{tabular} 
  \caption{\label{tab:different-models}Rouge F1 score of different models. We also report the results ($\dagger$) of several non-neural methods cited from \citet{fabbri2019multi}.}\label{tab:overall_performance}
  \end{table}

\subsection{Implementation Details}\label{sec:4.3}
We choose “bert-base-uncased” as our pre-trained BERT. For the NTM, we set the topic number $K$=50 and prune the vocabulary to 50,000. For the graph encoder, we set its layer number to 3. The dimension size of nodes representations is set to 768. For the decoder, we set the head of attention number to 6. $\gamma$ and $\tau$  is set to 0.8 and 0.3 to balance different losses. We train our model for up to 1000 epoch with a small batch size of 8. The experiments are based on 2 NVIDIA V100 cards. During the training, an early stop strategy is applied when the loss on validation set does not decrease for three consecutive epochs. We select the hyperparameters with grid search based on the Rouge-2 score on the validation set. In the summary generation, we adopt the beam search strategy with a search size of 5. We report the average results on 3 runs.

\section{Results and Analysis}\label{sec:5}

\begin{figure}[tp]
  \centering
  \includegraphics[scale=0.45]{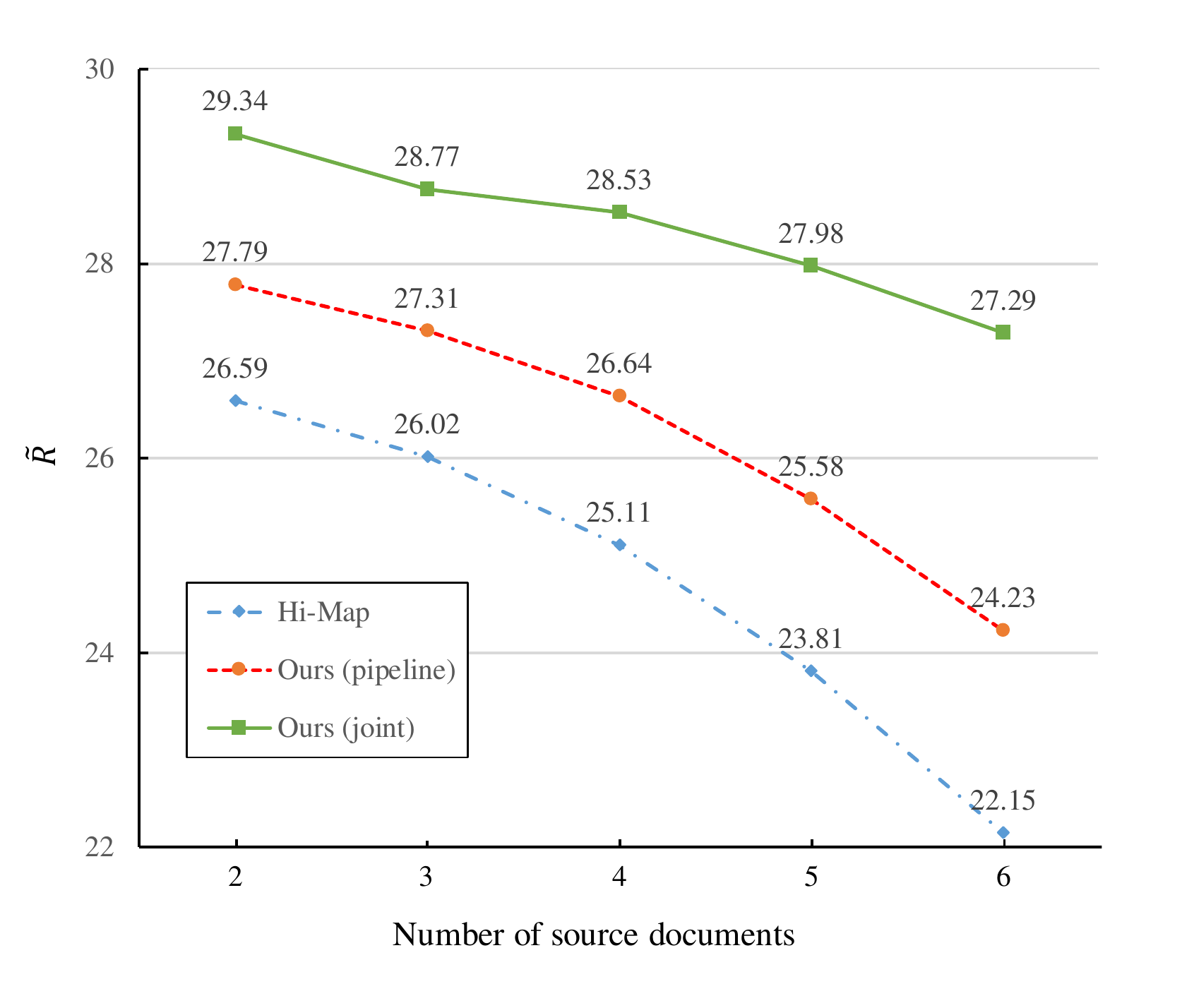}
 \caption{Relationship between number of source documents (x-axis) and model performance $\tilde{R}$ (y-axis), which is the mean of R-1, R-2, and R-SU.}\label{fig:doc_num}
\end{figure}   

\subsection{Automatic Evaluation}\label{sec:5.1}
\textbf{Overall Performance} \quad Table \ref{tab:overall_performance} presents the performance (Rouge) of our model against recently released methods on Multi-News. 
For ours, TG-MultiSum represents our jointly trained model, while TG-MultiSum${\rm _{pip}}$ is a pipeline version that separately trains the NTM and the summarizer.
We use it to verify whether joint topic inference can bring positive effect on summarization. 

As shown, our pipeline version shows competitive results against strong baselines and our full model achieves state-of-the-art performance on the Multi-News dataset, indicating topical information is an effective feature for summarization. 
Compared to TG-MultiSum$_{pip}$, our full model achieves 1.06/1.12/0.91 improvements on R-1, R-2, and R-SU.
This proves that joint topic inference is effective for abstractive MDS. 
We also observe that several graph-based models, such as MGSum and HeterGraphSum, achieve promising results compared to the "flat" models, such as PGN, and Copy-Transformer, implying that graph structure is an effective way to model multiple documents for MDS task.
Among the non-neural methods, Lead-3 serves as a simple but effective method. 
This is because that news articles tend to present key points in the beginning.

\noindent \textbf{Results on varying document numbers} \quad We also investigate how the source document number influences the model performance. To this end, we first divide the test set of Multi-News into different intervals based on the number of source documents and discard those with less than 100 examples. Then, we take Hi-Map\footnote[2]{We obtain similar results from other abstractive baselines.} as the baseline and compare the results on different parts. 

As shown in Figure \ref{fig:doc_num}, the Rouge declines with the increasing of document number, indicating that summarizing multiple documents is more challenging. Nevertheless, our two models show better robustness than Hi-Map on increasing document numbers. And joint training can further enhance this ability. Such observation verifies our assumption that latent topics can act as relay nodes to help capture cross-document relations for MDS.

\begin{figure}[tp]
  \centering
  \includegraphics[scale=0.45]{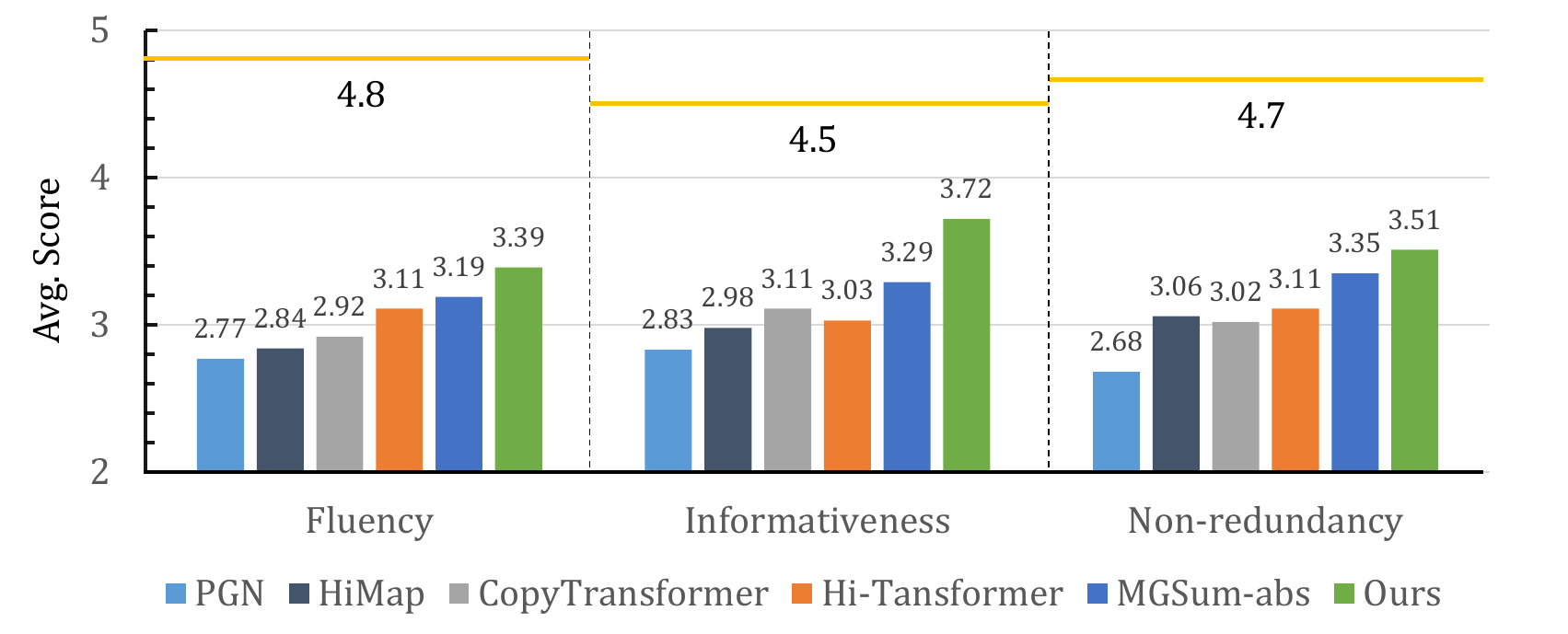}
  \caption{Human evaluation results of different abstractive models. Golden lines represent the scores of reference summary}
  \label{fig:human_evaluation}
\end{figure}

\subsection{Human Evaluation}\label{sec:5.2}
To evaluate the linguistic quality of generated summaries in better granularity, we conduct a human evaluation for the abstractive models based on three aspects: (1) \textbf{Fluency} measures whether the summary is coherent and grammatically correct. (2) \textbf{Informativeness} focuses on whether the summary covers the salient information of original documents. (3) \textbf{Non-redundancy} reflects whether the summary avoids repeated expressions. 
We sample 100 instances from the test set and generate summaries using different models. Then, we employ five graduates to rate the generated summaries.

As shown in Figure \ref{fig:human_evaluation}, our model beats all baselines in three indicators, especially in informativeness, implying that latent topics are indicative features for capturing salient information. 
Surprisingly, our model also shows promising improvement in non-redundancy score. 
This positive effect is probably attributed to the topic context $c_{*}^{T}$ (Eq.\ref{eq:9}) learned in the decoder. 
It can adaptively decide the current topic focus based on previous decoded words and therefore avoid generating repetitive contents of the same topic.  

We also present the human ratings of reference summaries (golden lines). As can be seen, despite the promising improvements of our model, there is a large gap between the quality of model-generated summaries and reference summaries, implying that abstractive MDS remains a challenge. 

\subsection{Ablation Study}\label{sec:5.3}
To analyze the relative contributions of different components to the model performance, we compare our full model with the following ablated variants: (1) \underline{w/o $\mathcal{L}_{inc}$} removes the inconsistency loss (Eq.\ref{eq:18}). (2) \underline{w/o topic nodes} builds the document graph solely with word and document nodes. (3) \underline{w/o topic pointer} removes the topic pointer (Eq.\ref{eq:9}) in the decoder. (4) \underline{w/o DGE} removes the document graph encoder described in \cref{sec:3.3}. (5) \underline{w/o NTM} removes the NTM module described in \cref{sec:3.1}. For compensation, we use a pre-trained LDA to provide word-topic matrix. (6) \underline{w/o BERT} removes the BERT encoder and initialize word and document nodes with trainable embeddings.

\begin{table}[tp]
  \small
  \centering
  \begin{tabular}{l|ccc}
  \toprule[1pt]
  \textbf{Ablated Models} & \multicolumn{1}{c}{\textbf{R-1}} & \multicolumn{1}{c}{\textbf{R-2}} & \multicolumn{1}{c}{\textbf{R-SU}} \\ \hline
  Our full model         & \multicolumn{1}{c}{\textbf{47.10}}        & \multicolumn{1}{c}{\textbf{17.55}}      & \multicolumn{1}{c}{\textbf{20.73}}     \\
  w/o $\mathcal{L}_{inc}$         & \multicolumn{1}{c}{46.69}        & \multicolumn{1}{c}{17.04}      & \multicolumn{1}{c}{20.15}       \\ 
  w/o topic nodes         & \multicolumn{1}{c}{46.48}        & \multicolumn{1}{c}{17.11}      & \multicolumn{1}{c}{20.08}       \\ 
  w/o topic pointer         & \multicolumn{1}{c}{46.32}        & \multicolumn{1}{c}{16.97}      & \multicolumn{1}{c}{19.73}      \\ 
  w/o DGE                & \multicolumn{1}{c}{46.08}        & \multicolumn{1}{c}{16.19}      & \multicolumn{1}{c}{19.72}      \\ 
  w/o NTM                  & \multicolumn{1}{c}{45.83}        & \multicolumn{1}{c}{16.02}        & \multicolumn{1}{c}{19.45}      \\ 
  w/o BERT                         & \multicolumn{1}{c}{45.67}        & \multicolumn{1}{c}{16.13}        & \multicolumn{1}{c}{19.27}       \\
  \bottomrule[1pt]   
  \end{tabular}
  \caption{Performance of different ablated variants against our full model) compared with our full model.} \label{tab:ablation_study}
  \end{table}

From Table \ref{tab:ablation_study}, We can obtain the following observations: 
(1) The removal of topic nodes and topic pointer both lead to performance drops, indicating that latent topics are effective features for both encoding and decoding process. 
(2) The document graph encoder plays a necessary role in our model since it can aggregate information from different granularities and documents. 
(3) NTM serves as a better topic learner than LDA in our experiments, and the inconsistency loss demonstrates its effectiveness. We conjecture that NTM can adaptively learn topics that are suitable for summarization under a multi-task setting where $\mathcal{L}_{inc}$ is applied, while the topics learned by LDA keep unchanged as external features \citep{cui2020enhancing}.
(4) The performance declines dramatically when removing BERT. This shows that BERT can provide necessary contextual information to better initialize the graph. Similar results have been observed in GNN-based extractive summarization \citep{wang2020heterogeneous}. 

\subsection{Impact of Hyperparameters}\label{sec:5.4}
\begin{figure}[tp]
  \includegraphics[scale=0.4]{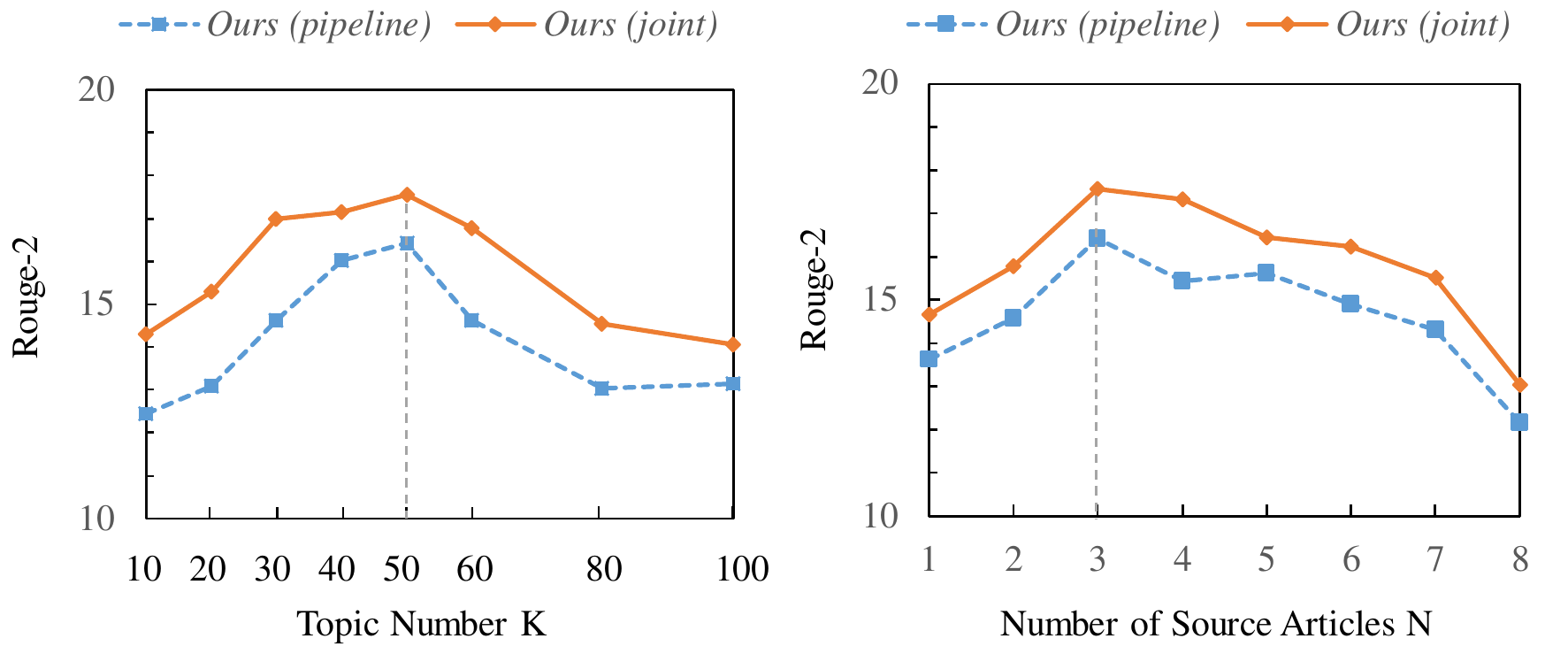}
  \caption{Impact of topic number (left) and graph iteration number (right) on model performance (Rouge-2).}\label{fig:parameters}
\end{figure}

We further conduct experiments on the validation set of Multi-News to probe into the impact of two important parameters, i.e., the topic number $K$ and the graph iteration number $L$. The results are presented in Figure \ref{fig:parameters}.

\noindent \textbf{Impact of topic number} \quad As can be seen, with a particular range where $K$ is relatively small, more topics produce better performance. However, such increasing trend will reach a saturation when $K$ exceeds a threshold (50 in our experiments). It worth noting that the joint model consistently beat the pipeline model under all different $K$, implying that joint training can help the NTM adaptively adjust learned topics for better summarization.

\noindent \textbf{Impact of graph iteration number} \quad Figure \ref{fig:parameters} (right) presents the relationship between graph iteration number with model performance. We can see that the two curves show a similar trend. 
In particular, the performance is dramatically boosted when $L$ goes from 0 to 3. 
However, such increasing trend is not always monotonous, and a larger $L$ will damage the performance. 
A possible reason lies in that deep networks could lead to overfit, although we add a residual connection between adjacent layers.

\subsection{Topic Quality Analysis}\label{sec:5.5}
We have shown the effect of latent topics on MDS task. In this subsection, we conduct experiments whether summary generation can in turn help in producing better topics.

We refer \underline{${\rm NTM_{sum}}$} as our jointly trained topic model and consider three baselines for comparisons. (1) \underline{${\rm LDA}$} \citep{blei2003latent} is a widely used topic model based on Bayesian graphical models. (2) \underline{BTM} \citep{yan2013a} is an enhanced topic model for short text modeling. (3) \underline{GSM} \citep{miao2017discovering} is the model used in our method. Different with ${\rm NTM_{sum}}$, it is separately trained on VAE loss. We use it to show the effect of joint summary generation on topic modeling.

The three comparison models are all trained on the Multi-News dataset. We run 1,000 Gibbs sampling for LDA and BTM to ensure the convergence. For GSM, we use the same settings described in \cref{sec:4.3} to make the results comparable.

\begin{table}[tp]
  \centering
  \small
  \begin{tabular}{c|c|c}
  \toprule
  \textbf{Models}       & $C_v$  & Sample Topics\\ \hline
  LDA       & \makecell{0.442} & \makecell[l]{sport NBA {\color{red} \underline{green}} champion watch \\
  deal guard brand {\color{red} \underline{speak}} commercial} \\ \hline
  BTM       & \makecell{0.431} & \makecell[l]{balls sport fight basketball {\color{red} \underline{year}} \\ {\color{red} \underline{violence}} foul superbowl fail {\color{red} \underline{crazy}}}\\ \hline
  GSM       & \makecell{0.370} & \makecell[l]{player sport {\color{red} \underline{eye}} football national \\  {\color{red} \underline{word}} halftime {\color{red} \underline{answer}} playing {\color{red} \underline{day}}}
  \\ \hline
  NTM$_{{\rm sum}}$ & \makecell{\textbf{0.496}} & \makecell[l]{sport quarterback scores NBA show \\ play reporter winner Olympic medal} \\
  \bottomrule
  \end{tabular}
  \caption{Coherence score $C_v$ and inferred topic (\emph{sport}) of different topic models. {\color{red} \underline{Off-topic words}} are underlined and in red.} \label{tab:topics}
\end{table}

\noindent \textbf{Topic Coherence} \quad Following previous studies \citep{zeng2018topic,wang2019topic}, we use the coherence score $C_v$ \citep{roder2015exploring}
to quantitatively evaluate inferred topics, which has been proved highly consistent with human evaluation. We can see from Table \ref{tab:topics} that the separately trained GSM performs rather poorly compared with two traditional models. However, the performance is significantly improved when it is jointly trained with the summarizer. This proves that a joint summarization task can in turn help in topic modeling because a summary usually reflects the major topics of its source document(s).

\noindent \textbf{Sample Topics} \quad To obtain a more intuitive comparison of the topic quality learned by different models, we present top 10 representative words of the topic "\emph{sport}" inferred by different models. As can be seen from Table \ref{tab:topics}, there are mixed off-topic words in three baselines. Besides, compared with them, our inferred topic looks more coherent. 
For example, it includes less half-related words, such as "\emph{commercial}" (LDA), "\emph{fail}" (BTM), and "\emph{national}" (GSM). 

\section{Conclusion and Future Work}
This study proposes a novel abstractive MDS model that integrates a joint NTM to discover latent topics.
Experimental results demonstrate that our model not only achieves the-state-of-the art results on summarization but also produce high-quality topics. 
Further discussions show that topic inference and summary generation can promote each other.
In the future, we will explore how to apply latent topics in controllable summarization.
\bibliography{custom}
\bibliographystyle{acl_natbib}

\end{document}